\documentclass[journal]{IEEEtran}
\usepackage{graphicx}
\usepackage{xcolor}
\usepackage{subcaption}
\usepackage{booktabs}
\usepackage{amssymb}
\usepackage{amsmath}
\graphicspath{{./images/}}
\usepackage{pgf}
\usepackage{hyperref}

\usepackage[absolute,showboxes]{textpos}
\TPMargin*{3pt}\newcommand\copyrighttext{    \footnotesize    \noindent    (c) IEEE 2024.\\    Personal use of this material is permitted.    Permission must be obtained for all other uses, in any current or future media, including reprinting/republishing this material for advertising or promotional purposes, creating new collective works, for resale or redistribution to servers or lists, or reuse of any copyrighted component of this work in other works.}
\newcommand\copyrightnotice{
    \begin{textblock*}{7in}(0.75in,0.15in)
    \copyrighttext
    \end{textblock*}
}

\newcommand{\padfigure}[1]{\vspace{-0.4cm}}

\DeclareMathOperator*{\argmin}{argmin}

\ifCLASSOPTIONcompsoc
  \usepackage[nocompress]{cite}
\else
  \usepackage{cite}
\fi

\begin{document}

\title{The Integration of Prediction and Planning in Deep Learning Automated Driving Systems: A Review}
\author{Steffen~Hagedorn$^{\ast 1,2}$ \quad Marcel Hallgarten$^{\ast 1,3}$ \quad Martin Stoll$^{1}$ \quad
Alexandru~Paul~Condurache$^{1,2}$\\
 $^1$Robert Bosch GmbH \quad $^2$University of Lübeck \quad$^3$University of Tübingen\\
 {\small $^\ast$These authors contributed equally to this work \quad Email: \{first-name\}.\{last-name\}@de.bosch.com \\}
}

\markboth{}
{Hagedorn, Hallgarten \MakeLowercase{\textit{et al.}}: The Integration of Prediction and Planning in Deep Learning Automated Driving Systems: A Review}

\maketitle
\begin{abstract}
Automated driving has the potential to revolutionize personal, public, and freight mobility.
Beside accurately perceiving the environment, automated vehicles must plan a safe, comfortable, and efficient motion trajectory.
To promote safety and progress, many works rely on modules that predict the future motion of surrounding traffic.
Modular automated driving systems commonly handle prediction and planning as sequential, separate tasks.
While this accounts for the influence of surrounding traffic on the ego vehicle, it fails to anticipate the reactions of traffic participants to the ego vehicle's behavior.
Recent methods increasingly integrate prediction and planning in a joint or interdependent step to model bidirectional interactions.
To date, a comprehensive overview of different integration principles is lacking.
We systematically review state-of-the-art deep learning-based planning systems, and focus on how they integrate prediction. 
Different facets of the integration ranging from system architecture to high-level behavioral aspects are considered and related to each other.
Moreover, we discuss the implications, strengths, and limitations of different integration principles.
By pointing out research gaps, describing relevant future challenges, and highlighting trends in the research field, we identify promising directions for future research.
\end{abstract}

\definecolor{signred}{HTML}{B85450}
{
\color{signred}
\textbf{Note:} A beginner-friendly introduction to prediction, planning, and their integration in automated driving systems can be found in an earlier version of our work at {\color{magenta}\url{https://arxiv.org/abs/2308.05731v2}}.
The following version is intended for more experienced researchers in the field. \\
}


\IEEEpeerreviewmaketitle
\copyrightnotice

\section{Introduction}\label{sec:introduction}
\begin{figure}
\centering
\includegraphics[width=\columnwidth]{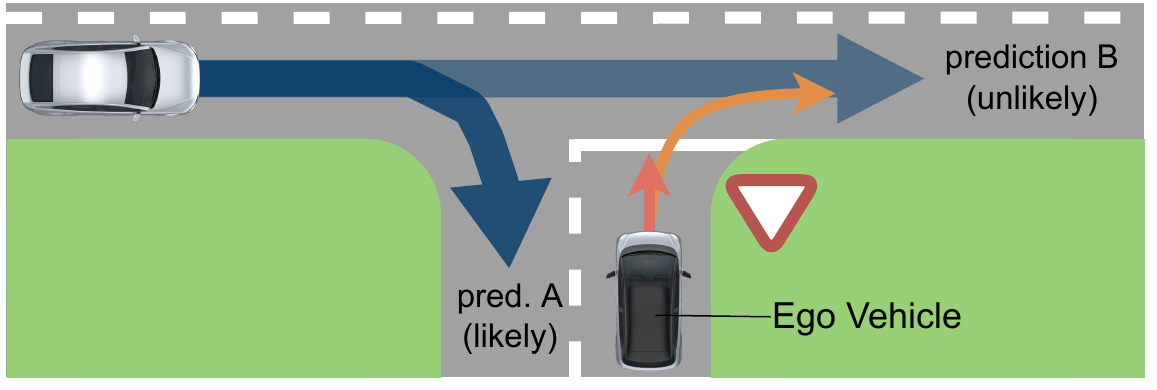}
\caption{\textbf{How should the ego vehicle behave?} Various behaviors are possible for the ego vehicle. Here, we depict two in red and orange. Their consequences depend on the behavior of the observed surrounding vehicle, which is depicted by blue arrows (prediction A or prediction B).
At the same time, the surrounding vehicle might react to the ego vehicle's action, i.e. the surrounding vehicle's behavior also depends on the ego vehicle's decision.
Different methods exist to forecast the behavior of the surrounding vehicle, and various ways exist to leverage this to decide on a safe and goal-directed plan for the ego-vehicle.
In this survey, we systematically categorize and review methods that integrate prediction and planning for self-driving vehicles.
We highlight their capabilities and limitations and discuss prospects for future research.
}
\label{fig:teaser_figure}
\padfigure{}
\end{figure}

\thispagestyle{empty}
\IEEEPARstart{A}{utomated} Driving (AD) remains a challenging endeavour. It is usually split into the subtasks of perception, prediction, planning, and control~\cite{yurtsever2020survey, gruyer2017perception, wen2022deep}.
Perception processes sensor inputs to create a model of the environment.
Prediction and planning build upon this model and make future motion forecasts for surrounding traffic agents and a plan for the controlled ego vehicle.
Traditional, modular systems (cf. Fig.~\ref{fig:overview_ads_ipp}) address prediction and planning as separate tasks. The predicted behavior of surrounding traffic agents is leveraged to plan a suitable behavior for the ego vehicle. 
However, this sequential ordering is inherently reactive and cannot represent bidirectional interaction between the ego vehicle and other traffic agents~\cite{sun2022m2i}.
In fact, prediction and planning are no sequential problems and should be tightly coupled in automated driving systems~\cite{rhinehart2021contingencies, ngiam2022scene, huang2022differentiable}. Fig.~\ref{fig:teaser_figure} highlights the interdependence of both tasks.

\subsection{Scope}
In this work, we review methods to integrate prediction and planning into an automated driving system.
Prediction is the task of anticipating intents and future trajectories of observed traffic agents~\cite{lee2017desire}, and planning is about finding the best possible trajectory w.r.t. previously defined criteria for a controlled vehicle~\cite{pomerleau1988alvinn}.
Deep Learning (DL)-based methods follow a data-centric approach to tackle these problems~\cite{wang2019deep} and have led to significant improvements in many fields~\cite{dong2021survey}. Works in prediction and planning increasingly employ DL-based approaches as well~\cite{grigorescu2020survey, kuutti2020survey, huang2020survey, huang2022survey} and represent an alternative to traditional, rule-based methods~\cite{treiber2000congested, kesting2007general,shalev2017formal} and constrained optimization~\cite{liu2017path}. In this work, we focus on DL-based methods.

\begin{figure*}
  \centering
  \includegraphics[width=\textwidth]{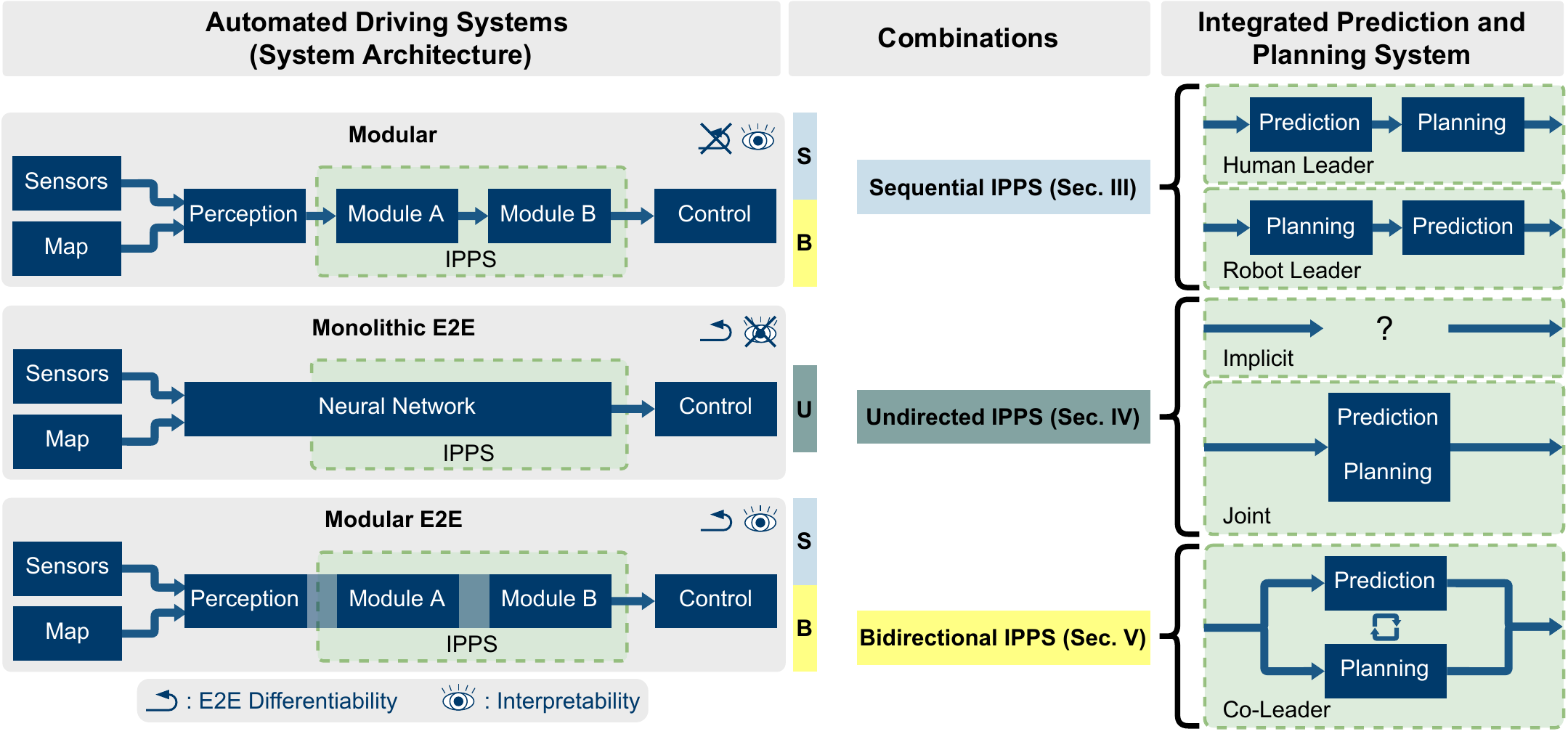}
  \caption{Overview of automated driving systems. There are three system architectures shown on the left. Modular systems consist of individual modules, whose interfaces provide interpretability but restrict information flow and end-to-end differentiability. In contrast, monolithic E2E systems are end-to-end differentiable but not interpretable. Modular E2E systems combine both properties: they are end-to-end differentiable and interpretable. All three system architectures comprise an integrated prediction and planning system (IPPS), depicted in green. Our work focuses on this very part and identifies three paradigms to integrate prediction and planning, as shown on the right. Sequential IPPSs condition one task on the other. Undirected IPPSs allow for more complex interactions but provide low interpretability. Bidirectional IPPSs explicitly ensure that both tasks are mutually conditioned on each other. Our review analyzes these integration paradigms and highlights their compatibility with different system architectures.}
  \label{fig:overview_ads_ipp}
\end{figure*}

An overview of different architectures of automated driving systems is shown in Fig.~\ref{fig:overview_ads_ipp}. It can take three forms ranging from modular systems over end-to-end (E2E) differentiable modular systems to monolithic E2E systems. 
Modular and modular E2E systems consist of clearly defined components for different tasks~\cite{hu2022st,muller2018driving, hawke2020urban, scheel2022urban, li2023planning}. 
Traditionally, they are arranged sequentially. We will discuss these systems in Sec.~\ref{sec:sequential_ipps}.
In contrast, monolithic E2E methods employ a single neural network for perception, prediction, and planning so that depending on the architecture, no boundaries between the modules can be drawn~\cite{pomerleau1988alvinn, bojarski2016end, chitta2022transfuser}.
As the order of the subtasks cannot be determined, we name this integration of prediction and planning ``undirected'' (c.f. Sec.~\ref{sec:undirected_ipps}).
In this work, we emphasize the shortcomings of sequential and undirected approaches and highlight works that integrate prediction and planning in a more sophisticated way to enable bidirectional interaction of self-driving vehicles and surrounding traffic (c.f. Sec.~\ref{sec:bidirectional_ipps}).
In the following, we refer to the component of an automated driving system that integrates prediction and planning as an Integrated Prediction and Planning System (IPPS).

\subsection{Contribution and Outline}
By discussing automated driving systems with a focus on the integration of prediction and planning, we take a novel perspective.
IPPSs can be classified into three categories as depicted in Fig.~\ref{fig:overview_ads_ipp}.
Accordingly, this survey is structured into three main chapters: sequential (Sec.~\ref{sec:sequential_ipps}), undirected (Sec.~\ref{sec:undirected_ipps}), and bidirectional IPPSs (Sec.~\ref{sec:bidirectional_ipps}).
While surveys on classical methods~\cite{lefevre2014survey, claussmann2019review, huang2020survey, leon2021review}, stand-alone DL-based prediction~\cite{mozaffari2020deep, liu2021survey, ding2023incorporating, huang2022survey} or planning~\cite{schwarting2018planning}, and E2E AD~\cite{chen2023end} already exist, we observe that recent methods thoughtfully design the interplay of prediction and planning~\cite{huang2023gameformer, jiang2023vad, hu2023planning, huang2023dtpp}. To substantiate this observation with broad evidence and a theoretical foundation, we summarize our contributions as follows:

\begin{itemize}
    \item We propose a categorization for the integration of prediction and planning based on the dependencies between both tasks (cf. Sec.~\ref{sec:sequential_ipps},~\ref{sec:undirected_ipps} and~\ref{sec:bidirectional_ipps}). We investigate how these categories relate to system architectures, behavioral aspects, and safety.
    \item In particular, we provide a comprehensive overview of design choices in prediction and planning modules of modular (E2E) architectures (cf. Sec.~\ref{sec:sequential_ipps_prediction} and~\ref{sec:sequential_ipps_planning}) and discuss their impact on interaction and system-level behavior.
    \item We reveal gaps in state-of-the-art research and point out promising directions of future research based on the identified categorization (Sec.~\ref{sec:discussion_frontiers}).
\end{itemize}

\hfill

\section{Task Definitions}\label{sec:task_definitions}
In the following, we start by introducing the terminology and notation for the task definitions used in this work.
We adopt a similar terminology to that proposed by~\cite{mozaffari2020deep} and partition the actors in a traffic scenario into the self-driving ego vehicle (EV) and the surrounding vehicles (SVs). 
The EV is equipped with sensors that provide information on the environment.
The state history of vehicle $i$ over the time interval $t-t_\textrm{obs}, \ldots, t$ is
\begin{equation}
    X_i=\{x_{t-t_\textrm{obs}}, \ldots, x_{t-1}, x_{t}\}.
\end{equation}
Each state $x$ comprises 2D or 3D positional information and further optional information like heading angle, speed, static attributes, or goal information in the case of the EV.
Hence, $X_\textrm{EV}$ denotes the EV's past states. Similarly,  
\begin{equation} \label{eqn:past_states_x}
    \overline{\textbf{X}}_\textrm{SV} = \{ X_1, X_2, \ldots, X_m \}
\end{equation}
refers to the past states of all SVs.
Analogously, the future states of vehicle $i$ within a prediction horizon of $t_\textrm{pred}$ are
\begin{equation}
    Y_i=\{y_{t+1}, y_{t+2}, \ldots, y_{t+t_\textrm{pred}}\}.
\end{equation}
and the future states of all SVs are $\overline{\textbf{Y}}_\textrm{SV}$.
In the following, state sequences are also referred to as trajectories.
Additional scene information, such as a semantic map or traffic signs and traffic light states are represented by $I$.

Following~\cite{lee2017desire}, we state trajectory \textit{prediction} as the task of estimating a probability distribution
\begin{equation}\label{eqn:traj_forecasting}
    P_\textrm{pred} = P(\overline{\textbf{Y}}|\overline{\textbf{X}}, I)
\end{equation}
that maps state histories of $m$ observed vehicles in $\overline{\textbf{X}}$ to the future trajectories of $n$ predicted vehicles in $\overline{\textbf{Y}}$.
The distribution $P_\textrm{pred}$ accounts for inherent uncertainties in the forecasting task and is often modeled by a discrete set of samples with corresponding probabilities~\cite{varadarajan2022multipath++,deo2020trajectory, salzmann2020trajectron++, liang2020learning}.
Some methods omit scene information $I$ and infer trajectories based on a state history alone~\cite{mercat2020multi, xu2022trajectory, schmidt2022crat}.
Commonly $\overline{\textbf{X}}$ includes all vehicles in the scene, i.e. $X_\textrm{EV}$ and $\overline{\textbf{X}}_\textrm{SV}$. Depending on the vehicles in $\overline{\textbf{Y}}$, different variants of prediction can be formulated:
single-agent prediction models the future trajectory for each SV individually~\cite{bhattacharyya2024ssl}.
However, estimating a joint prediction $\overline{\textbf{Y}}_\textrm{SV}$ from a set of $n$ single-agent predictions $Y_\textrm{SV}$ is not trivial since the number of possible combinations grows exponentially with the number of agents $n$.
Not all combinations are meaningful, and finding realistic ones with heuristics and joint optimization is cumbersome~\cite{gilles2021thomas}.
To avoid this, joint prediction directly estimates a joint distribution for multiple SVs~\cite{casas2020implicit}.

In \textit{planning}, the task is to find a single suitable trajectory for the EV that can be passed on to a downstream motion controller. Hence, we define planning to be a function $f$ that maps the observational inputs $\overline{\textbf{X}}_\textrm{SV}$, and $X_\textrm{EV}$ as well as the context information $I$ to a future trajectory $Y_\textrm{EV}$, i.e.
\begin{equation}\label{eqn:definition_planning}
    Y_\textrm{EV}=f(X_\textrm{EV}, \overline{\textbf{X}}_\textrm{SV}, I).
\end{equation}
In many cases, the $f$ also uses the prediction $\overline{\textbf{Y}}_\textrm{SV}$, i.e.
\begin{equation}\label{eqn:definition_planning_extended}
    Y_\textrm{EV}=f(X_\textrm{EV}, \overline{\textbf{X}}_\textrm{SV}, I, \overline{\textbf{Y}}_\textrm{SV}).
\end{equation}
The definitions show that planning can be considered a special case of single-agent prediction where the output distribution models only a single trajectory.
However, in contrast to prediction, the planning trajectory must be conditioned on a navigation goal.
Moreover, it has to be stable in closed-loop, i.e. it must be kinematically feasible and result in safe and efficient driving behavior when fed to a downstream controller~\cite{rajamani2011vehicle}.

\section{Sequential IPPS}\label{sec:sequential_ipps}
To date, the sequential integration of prediction and planning is the prevalent and most widely adopted integration principle.
Sequential IPPSs execute prediction and planning as separate tasks.
Prior knowledge is used to design the interplay of the modules.
Depending on the interfaces, such systems can still be end-to-end differentiable.
Accordingly, sequential integration can be found in modular and modular E2E systems (c.f. Fig.~\ref{fig:overview_ads_ipp}). 
To investigate different sequential system designs in more depth, we start by taking a closer look at the individual components of prediction and planning and discuss their impact on interaction and system-level behavior.

\subsection{Prediction}
\label{sec:sequential_ipps_prediction}
\subsubsection{Input Representations}
\label{sec:pred_scene_repr}

In the context of IPPSs for automated vehicles, agent states $\overline{\textbf{X}}$ and map $\textit{I}$ are the most important information to represent. 
Two major representations exist in DL-based approaches: Rasterized and sparse.

\textit{Rasterized} representations use dense, fixed-resolution grid structures that often have multiple channels~\cite{casas2018intentnet, park2020diverse, chen2022scept, kamenev2022predictionnet, stoll2023scaling, janjos2021self}.
Each channel encodes different information on agent states.
They are commonly used in combination with rasterized high-definition maps (HD maps)~\cite{bansal2018chauffeurnet, mi2021hdmapgen} and represented in a bird's eye view (BEV)~\cite{philion2020lift, li2022bevformer}.
The BEV allows to fuse multiple EV sensor inputs and different sensing modalities into a shared and interpretable representation~\cite{lee2017desire}.
It further establishes a common coordinate system for all vehicles, facilitating to model interactions within the scene~\cite{djuric2018short}.
Dense grids are well-suited for processing with powerful Convolutional Neural Network (CNN)~\cite{lecun1989handwritten} architectures~\cite{bansal2018chauffeurnet,zeng2019end,rhinehart2018deep,wang2019deep,chen2019deep,codevilla2019exploring,rhinehart2019precog, sadat2020perceive,chen2020learning,zeng2020dsdnet,cui2021lookout,casas2021mp3,konev2022motioncnn}.
However, embedding all observations in a grid structure leads to information loss due to quantization errors~\cite{diehl2019graph}. 
Moreover, the locally restricted receptive field and the limited resolution of CNNs can hamper interaction modeling.

\textit{Sparse} representations use vectors to describe all objects in a scene. The per-object vectors are then either encoded into a latent representation jointly, e.g., with a Transformer, or individually and then aggregated, e.g., with Graph Neural Networks (GNN)~\cite{kipf2016semi}. 
Objects are represented by polygons, or point sets~\cite{ngiam2022scene, gao2020vectornet, liang2020learning, zhao2021tnt, zeng2019end, liu2021multimodal, varadarajan2022multipath++, mu2024most, deo2022multimodal, janjos2022san, deo2020trajectory, mo2022multi, pittner2024lanecpp}. 
The map is commonly described by a set of lanes represented by polylines and an adjacency matrix.
Objects and map elements are then encoded into fixed-size latent features, for instance, via Multi-Layer Perceptrons (MLP)~\cite{rosenblatt1958perceptron} or Recurrent Neural Networks (RNN)~\cite{rumelhart1986learning}.
To encode sequential signals like state histories $\overline{\textbf{X}}$, 1D CNNs~\cite{kim2021lapred}, RNNs like LSTMs~\cite{hochreiter1997long} and GRUs~\cite{chung2014empirical}, and Transformers~\cite{vaswani2017attention} are applied~\cite{liang2020learning, gilles2022gohome, liu2021multimodal}.
Some works like Trajectron++~\cite{salzmann2020trajectron++} and MFP~\cite{tang2019multiple} combine rasterized HD maps with sparse object representations.

Throughout the past years, a clear trend towards sparse input representations can be observed in prediction modules~\cite{zhou2022hivt, nayakanti2022wayformer, shi2022motion, cui2022gorela, jiang2023motiondiffuser, liu2024laformer, philion2024trajeglish, liao2024bat}.
In the context of integrating prediction and planning, an advantage of sparse representations is that they are object-based.
Interactions between the EV and SVs can therefore be modeled in a targeted manner.
This improves the explainability of a system and provides a clear understanding of how surrounding vehicles influence the ego-plan.

\subsubsection{Interaction Modelling}
Provided with an accurate scene representation, interaction modeling between perceived objects is a decisive factor for the successful integration of prediction and planning~\cite{sun2022m2i, tolstaya2021identifying}.
Modeling interactions between vehicles helps to understand how the actions of a vehicle affect the behavior of other traffic participants~\cite{liao2024bat}.
This model is subject to various influences, including traffic rules, physics, and common sense reasoning.
Interaction modeling also includes interactions between map elements and vehicles.
For example, modeling the relation of vehicles to static map elements like lane markings, traffic signs, etc.\ is important to understand feasible corridors in which trajectories could be localized.
Some models employ rule-based heuristics to guide interaction modeling, e.g., ScePT~\cite{chen2022scept} only models interaction among nearby vehicles. Similarly, heterogeneous graph attention network (HGAT)~\cite{demmler2024towards} and graph-based interaction-aware trajectory prediction (GRIP)~\cite{li2019grip} use heuristics to identify which pairs of nodes in the scene graph are to be connected with edges.
Other methods rely entirely on neural components to learn these relations.
Altogether, an interaction model should yield an understanding of how the scene can potentially evolve in the future.

Recapitulating recent methods exposes that diverse architectures are used to realize interaction modeling.
\textit{RNNs} are only used in early prediction models like DESIRE~\cite{lee2017desire} or Trajectron++~\cite{salzmann2020trajectron++} and combined with an aggregation operator like spatial pooling or attention.
Alternatively, \textit{CNN} methods apply 2D convolutions to implicitly capture interaction within the kernel size~\cite{luo2018fast, cui2019multimodal, chai2019multipath}.
Compared to sequence processing-based approaches, higher importance is assigned to spatial interaction.
However, no explicit interactions between designated objects are modeled.
This shortcoming is addressed by \textit{GNNs and Graph-Attention}, which explicitly model interactions between individual agents.
They either combine multiple agents' features~\cite{zhang2024graphad} and process them with graph convolution operators~\cite{wang2019dynamic, kipf2016semi, li2019grip, li2019grip++} or perform graph attention~\cite{bahdanau2014neural, luong2015effective, velickovic2017graph, pan2020lane, gu2021densetnt} to aggregate information.
Recently, \textit{Transformers}~\cite{vaswani2017attention} are widely adopted for interaction modeling due to their global receptive field and attention mechanism~\cite{li2020end, liu2021multimodal, yuan2021agentformer, ngiam2022scene, zhou2022hivt, nayakanti2022wayformer, shi2022motion, jiang2023motiondiffuser, philion2024trajeglish, chen2021s2tnet, quintanar2021predicting, girgis2021latent, postnikov2021transformer, singh2022multi, hazard2022importance, zhang2022trajectory, wonsak2022multi, amirloo2022latentformer, hu2023holistic}.
All SVs can be predicted simultaneously~\cite{liu2021multimodal}, and the impact of vehicles on each others' behaviors across different timesteps can be modeled by masking~\cite{yuan2021agentformer}. 
Attention is applicable along the spatial and temporal dimension jointly~\cite{yuan2021agentformer}, or individually~\cite{chen2023categorical}.
The latter break down into sequential~\cite{arnab2021vivit, ho2019axial} and interleaved~\cite{ngiam2022scene, nayakanti2022wayformer} attention.
Experiments with Wayformer~\cite{nayakanti2022wayformer} and SceneTransformer~\cite{ngiam2022scene} show that joint and interleaved processing outperform sequential approaches. 
As joint processing causes a higher computational burden, the interleaved paradigm balances best between performance and computational demand.
This property of Transformers can provide explainability to the integration of prediction and planning. Attention can be utilized flexibly to achieve locality restrictions, causality, and an explicit interaction design.
For example, a sequential IPPS could first model interactions between SVs and then include the EV for planning.

\subsubsection{Coordinate Frame}
Irrespective of the scene representation itself, the coordinate frame is a fundamental design choice that affects interaction modeling and thus the integration of prediction and planning.
Applying a global coordinate system with a fixed viewpoint for the whole scene~\cite{khandelwal2020if, gilles2021thomas, zeng2021lanercnn, ngiam2022scene} is computationally efficient as it can be shared across both modules.
However, the frame is not viewpoint-invariant, i.e., the predictions and the plan change if the origin of the coordinate system is moved within the scene. 
This leads to low sample efficiency and impairs generalization~\cite{reiter2024equivariant}. 
Therefore, the origin is commonly fixed to the center of the EV.
While this normalization with respect to the EV is a natural choice when planning for the EV, it does not lead to viewpoint-invariant predictions for SVs~\cite{hagedorn2024pioneering}.
In addition, it complicates the modeling of interactions among SVs. 
For instance, this coordinate frame makes it easy to infer which vehicles are close to the EV or in front of it (small magnitude of $x$ position). However, finding SVs that are close to each other is a second-order relation that is harder to learn.
This is why other models~\cite{janjos2022starnet, jia2022multi, gao2020vectornet, cui2019multimodal} process a scene from the viewpoint of each agent. Thus, the processing is viewpoint-invariant and makes it easier to represent interactions between all agents. On the downside, these models cannot use a shared representation for prediction and planning, and computational complexity scales linearly with the number of agents~\cite{ngiam2022scene}.
Moreover, per-agent coordinate frames are unsuitable for joint prediction, as each agent is processed individually.

An alternative way to achieve viewpoint invariance is the pairwise relative coordinate system introduced in GoRela~\cite{cui2022gorela}, which describes the relation between agents instead of using a fixed coordinate frame.
This allows the computation of the relation between static objects offline and greatly reduces computation.
Pioneering Equivariant Planning (PEP)~\cite{hagedorn2024pioneering} further reduces computation by using center-relative instead of pairwise relative coordinates, i.e. shifting the coordinate frame into the mean position of all processed agents before each transformation.
This makes it easier to represent and model interactions. It also eases the integration of prediction and planning, as both can use a shared scene representation. Thus, they can be combined in a single compute-efficient step.

The downside of the Cartesian frames mentioned above is that the models are vulnerable to distributional shifts and often produce inadmissible offroad trajectories~\cite{bahari2022vehicle,hallgarten2023stay}. 
Some works counter this with loss functions that penalize offroad driving~\cite{ridel2020scene,messaoud2020multi,niedoba2019improving,boulton2020motion}, driving direction compliance~\cite{greer2021trajectory} or being distant from the centerline~\cite{casas2020importance}.

An alternative approach is to use Frenet representations that decompose positions into progress along and lateral offset from a lane~\cite{houenou2013vehicle, yao2013lane, gilles2022gohome, zhang2021map}.
While this improves map compliance (i.e., reduces offroad), it impairs interaction modeling due to the non-euclidean coordinate frame. Moreover, this coordinate frame cannot be shared across multiple agents.

\subsubsection{Output Representations}
\label{sec:pred_output_representations}
As suggested in Fig.~\ref{fig:overview_ads_ipp}, the interfaces between modules are crucial for the overall system performance.
To optimize an automated driving system for the final planning task, differentiable interfaces are required~\cite{hu2023planning, karkus2023diffstack}.
To provide interpretability, the interfaces should comprise a human-readable representation~\cite{zeng2019end}.
When predicting trajectories in modular E2E systems, the output must fulfill these two criteria.
Additional difficulty arises from the inherent uncertainty of future prediction~\cite{hubmann2018automated, chen2024vadv2}.
Since the intentions of SVs are unknown, multiple future modalities must be considered to make a safe plan~\cite{rhinehart2021contingencies}.
Therefore, the multimodality of $P_\textrm{pred}$ is commonly expressed as an explicit density function or estimated with a discrete trajectory set.

Explicit density functions model the probability density function of $P_\textrm{pred}$~\cite{zeng2019end, sadat2020perceive, cui2021lookout, casas2021mp3}.
Continuous distributions like bi-variate Gaussian distributions~\cite{luo2020probabilistic}, Gaussian Mixture Models~\cite{shi2022motion, knittel2023dipa, huang2023gameformer, liu2024laformer}, or discrete rasterized heatmaps~\cite{gilles2021home, gilles2022gohome} are used in many prediction modules.
Alternatively, object-agnostic representations like occupancy maps~\cite{kim2017probabilistic, kim2022stopnet}, or flow fields~\cite{mahjourian2022occupancy, agro2023implicit, casas2021mp3} can be used.
Object-agnostic representations handle multimodality implicitly~\cite{zeng2019end} and were shown to be more robust against perturbations~\cite{rhinehart2018deep}.
Further, density functions are interpretable for humans.
However, since no discrete trajectories are decoded, it is difficult to compare such representations with expert logs for performance evaluation.

In an alternative approach to explicit density functions, discrete trajectory sets are either created by sampling from an intermediate distribution or directly output by the model.
To sample a diverse trajectory set from an intermediate distribution, generative models such as Generative Adversarial Networks (GANs)~\cite{goodfellow2020generative}, Conditional Variational Autoencoders (CVAEs)~\cite{sohn2015learning}, normalizing flows~\cite{rezende2015variational} or denoising diffusion~\cite{jiang2023motiondiffuser}, can be used~\cite{lee2017desire, zhao2019multi, sriram2020smart, choi2021drogon, dendorfer2021mg}.
Alternatively, recent methods apply several prediction heads to the features extracted by a common backbone to decode a set of diverse trajectories~\cite{varadarajan2022multipath++,liang2020learning, wang2022tenet, nayakanti2022wayformer}.
Other ways to obtain diverse predictions are training loss functions~\cite{zhao2021tnt}, entropy maximization~\cite{deo2020trajectory, aghasadeghi2011maximum}, variance-based non-maximum suppression~\cite{wang2022ltp}, greedy goal sampling~\cite{phan2020covernet, cui2022gorela}, divide and conquer strategy~\cite{narayanan2021divide}, evenly spaced goal states~\cite{lu2022kemp}, ensembling~\cite{ye2022dcms}, or using pre-defined anchor trajectories~\cite{chai2019multipath, song2022learning, fang2020tpnet}. Nevertheless, a guarantee of coverage can only be given by specifying high-level behaviors~\cite{chen2022learning} or regions of the map that must be covered by at least one prediction~\cite{liu2021multimodal}.

\subsection{Planning}
\label{sec:sequential_ipps_planning}
\subsubsection{Input Representations}
The input of the planner depends on the order of modules in sequential IPPSs.
If prediction is conditioned on potential EV actions, planning is executed first and takes the sparse or rasterized scene representation (cf. Sec.~\ref{sec:pred_scene_repr}) as input.
If prediction is executed before planning, the predictions are fed to the planner as an additional input in one of the formats described in Sec.~\ref{sec:pred_output_representations}.

In both cases, goal information for the EV is available in the planning module.
It can have different forms like lane-level route information~\cite{caesar2021nuplan}, high-level commands~\cite{dosovitskiy2017carla, zhai2023rethinking}, and sparse goal points~\cite{dosovitskiy2017carla}.
There are four categories how goal information is incorporated into the IPPS: input features, separate submodules, routing cost, and route attention.
Simply providing goal information as input feature yields no guarantees for goal compliance but is straightforward and widely adopted~\cite{codevilla2018end, chitta2021neat, chitta2022transfuser, chen2022learning, shao2023safety, wu2022trajectory, scheel2022urban, chen2019deep, codevilla2019exploring, hecker2018end, hawke2020urban, ohn2020learning}.
Separate submodules are used only with high-level commands, which serve as a switch between command-specific submodules~\cite{codevilla2018end, muller2018driving, sauer2018conditional, chen2020learning, chen2022learning}.
This can prevent the model from over-adapting to driving modes that dominate the training dataset, but requires a fixed number of high-level commands to be defined in advance.
A third option to incorporate goal information is by optimizing a cost term, which comprises progress and route compliance~\cite{sadat2020perceive,cui2021lookout,rhinehart2018deep}.
Balancing multiple targets in a cost term provides flexibility and interpretability but can lead to undesirable tradeoffs with safety.
Lastly, route attention applies spatial attention to on-route parts of the map.
This is achieved by Transformer cross-attention~\cite{bronstein2022hierarchical} or by removing the off-route portion of the map after initial feature aggregation~\cite{dauner2023parting, hagedorn2024pioneering, hallgarten2023prediction, chen2023tree, huang2023dtpp}.

Since sequential IPPSs consist of discernible modules, the interfaces between them have to comply with diverse requirements.
On one hand they should be interpretable to understand failures and facilitate development.
Rasterized~\cite{bansal2018chauffeurnet,chen2020learning,chen2022learning,wang2019monocular} and sparse~\cite{renz2022plant,pini2022safe,hallgarten2023prediction,vitelli2022safetynet,scheel2022urban} scene representations (cf. Sec.~\ref{sec:pred_scene_repr}) are teherefore common intermediate representations, which are input to the planning module.
Similar to the input representation used by prediction modules, a clear trend can be observed from rasters to sparse inputs~\cite{bansal2018chauffeurnet, zeng2019end, huang2023gameformer, huang2023dtpp}.
On the other hand an interface should convey all relevant information to downstream modules to cover the long-tailed distribution of potential driving scenarios.
Irrespective of the order of modules, information loss is a problem of hand-crafted interfaces~\cite{karkus2023diffstack}.
For instance, if a vehicle is currently occluded, but its presence can be inferred from the reflection of its headlight in another car, bounding box representations~\cite{bansal2018chauffeurnet}, occupancy fields~\cite{casas2021mp3}, or affordances~\cite{chen2015deepdriving, sauer2018conditional} cannot propagate this information to downstream modules~\cite{karkus2023diffstack, hu2023planning}.
Conversely, latent feature representations allow uncertainty propagation, enabling downstream modules to compensate for errors at earlier stages, such as undetected vehicles~\cite{zeng2019end}.

\subsubsection{Planning Paradigms}
\label{sec:planning_paradigms}
When taking a closer look at the planning module, three general planning paradigms can be observed.
We introduce each paradigm and discuss its suitability for the integration of prediction and planning.
Recall that in Eqns.~\ref{eqn:definition_planning} and~\ref{eqn:definition_planning_extended} we defined planning to be a function $f$ mapping from observational inputs $X_\textrm{EV}$, $\overline{\textbf{X}}_\textrm{SV}$, $I$ and possibly predictions $\overline{\textbf{Y}}_\textrm{SV}$ to a trajectory $Y_\textrm{EV}$.
In the following, we focus on the planning function $f$.
To this end, we decompose $f$ into two parts:
a proposal generator $g$, yielding a set of multiple potentially suitable trajectories $\hat{Y}^{(i)}_\textrm{EV}, i=1,..,N_\textrm{proposals}$ and a proposal selector $h$ that selects the final plan $Y_\textrm{EV}$ among the proposed set.
Hence, in this work, we propose to represent the planning function $f$ as
\begin{equation}
\label{eqn:planning_fct_unconditioned}
    f=h(g(X_\textrm{EV},\overline{\textbf{X}}_\textrm{SV},I)
\end{equation}
or in case of prediction-conditioned planning as
\begin{equation}
\label{eqn:planning_fct_conditioned}
    f=h(g(X_\textrm{EV},\overline{\textbf{X}}_\textrm{SV},I,\overline{\textbf{Y}}_\textrm{SV})\,.
\end{equation}
Based on this, we distinguish three different paradigms adopted for the planning task: Cost function optimization, regression, and hybrid planning.

\textit{Cost function optimization.} 
As planning aims to find a trajectory that optimizes objectives such as safety, comfort, and progress, it is a natural approach to design and optimize a cost function that balances these potentially conflicting objectives.
Cost function-based planning methods~\cite{sadat2019jointly, zeng2019end, biswas2024quad} rely entirely on the selection function $h$ to find a suitable trajectory,
\begin{equation}
    h=\argmin_i c\left(\hat{Y}^{(i)}_\textrm{EV}\right),
\end{equation}
where $c$ is a cost function of the planned trajectory.
Consequently, the proposal generator $g$ may randomly sample feasible motion profiles~\cite{zeng2019end} or obtain them by clustering real-world expert demonstrations~\cite{casas2021mp3}.
Traditional non-learning-based methods rely on hand-crafted cost functions~\cite{truong2023paas, werling2010optimal, buehler2009darpa, fan2018baidu, montremerlo2008stanford, ziegler2014trajectory, ajanovic2018search, pulver2021pilot}.
However, designing a cost function that effectively balances these goals and generalizes to the long-tailed distribution of possible driving scenarios is challenging.
Learning-based methods aim to address this by learning a cost function directly from expert demonstrations~\cite{ziebart2008maximum, wulfmeier2015maximum, rhinehart2018r2p2}.
The learned cost function can be non-parametric~\cite{zeng2019end, wei2021perceive,hu2021safe} or comprise hand-designed sub-costs of which only the weights are learned~\cite{cui2021lookout,casas2021mp3,sadat2020perceive,sadat2019jointly}.
Cost function-based planning enables the intuitive integration with a prediction module by including predictions in the cost function~\cite{zeng2020dsdnet, cui2021lookout}.
Depending on the prediction output, this can be realized either by explicit collision checking~\cite{vitelli2022safetynet} or by adopting predicted density functions~\cite{zeng2019end} as non-parametric cost.
For example, the NMP~\cite{zeng2019end} predicts a cost volume, which assigns a cost to each potential future EV position within the planning horizon.
In contrast, P3 (Perceive, Predict, and Plan)~\cite{sadat2020perceive} and MP3 (Map, Perceive, Predict and Plan)~\cite{casas2021mp3} apply hand-designed sub-costs with learned weighting to heuristically evaluate safety, comfort, traffic rule compliance, and progress of candidate trajectories~\cite{sadat2019jointly}.
In both cases, the cost function is then used to select the min-cost plan from the candidates.

\textit{Regression} approaches rely entirely on the proposal generator $g$.
The selection function $h$ is the identity, and all burden is placed on the proposal generator $g$, which only generates a single proposal $Y_\textrm{EV}=\hat{Y}^{(1)}_\textrm{EV}$.
Typically, regression-based planning is not used in sequential IPPSs but rather found in undirected IPPSs~\cite{pomerleau1988alvinn,bojarski2016end,xu2017end,sauer2018conditional,hecker2018end,muller2018driving,rhinehart2018deep,wang2019deep,chen2019deep,codevilla2019exploring,hawke2020urban, chen2020learning,ohn2020learning,chitta2022transfuser,scheel2022urban,chen2022learning,renz2022plant}.
Unlike cost function-based planning, regression cannot include predictions explicitly.
One possibility is to condition the plan on predictions as stated in Eqn.~\ref{eqn:planning_fct_conditioned}.
Regression approaches directly output future waypoints ~\cite{hallgarten2023prediction, chitta2022transfuser, scheel2022urban} or, more rarely, actions~\cite{pomerleau1988alvinn,bojarski2016end,chen2015deepdriving,xu2017end}.
Both representations can be converted into one another using a kinematic model~\cite{rajamani2011vehicle, janjos2021self}.
Commonly, they are trained by supervised learning from expert data~\cite{schaal1996learning}, called behavior cloning~\cite{bain1995framework, ross2011reduction, daftry2017learning}. This is easy to implement but vulnerable to distributional shifts, which occur when the model reaches states not covered by the training data~\cite{scheel2022urban}.
Mitigation strategies include heuristically generated data augmentations or policy rollouts during training~\cite{scheel2022urban, kumar2022cw}.

\textit{Hybrid planning} describes methods that combine both ideas.
First, a set of candidate trajectories is regressed by the proposal generator $g$. Then the best one is selected with a cost function in $h$~\cite{cheng2024pluto}.
For instance,~\cite{vitelli2022safetynet, pini2022safe, hu2023imitation, yi2023imitation, huang2023gameformer} perform collision checks on the candidate plans to rule out unsafe proposals.
In contrast to cost function optimization, the generated trajectory proposals are not fixed but scene-dependent.

\subsection{Integration}
\subsubsection{Human Leader}
\begin{figure}
    \centering
    \includegraphics[width=\linewidth]{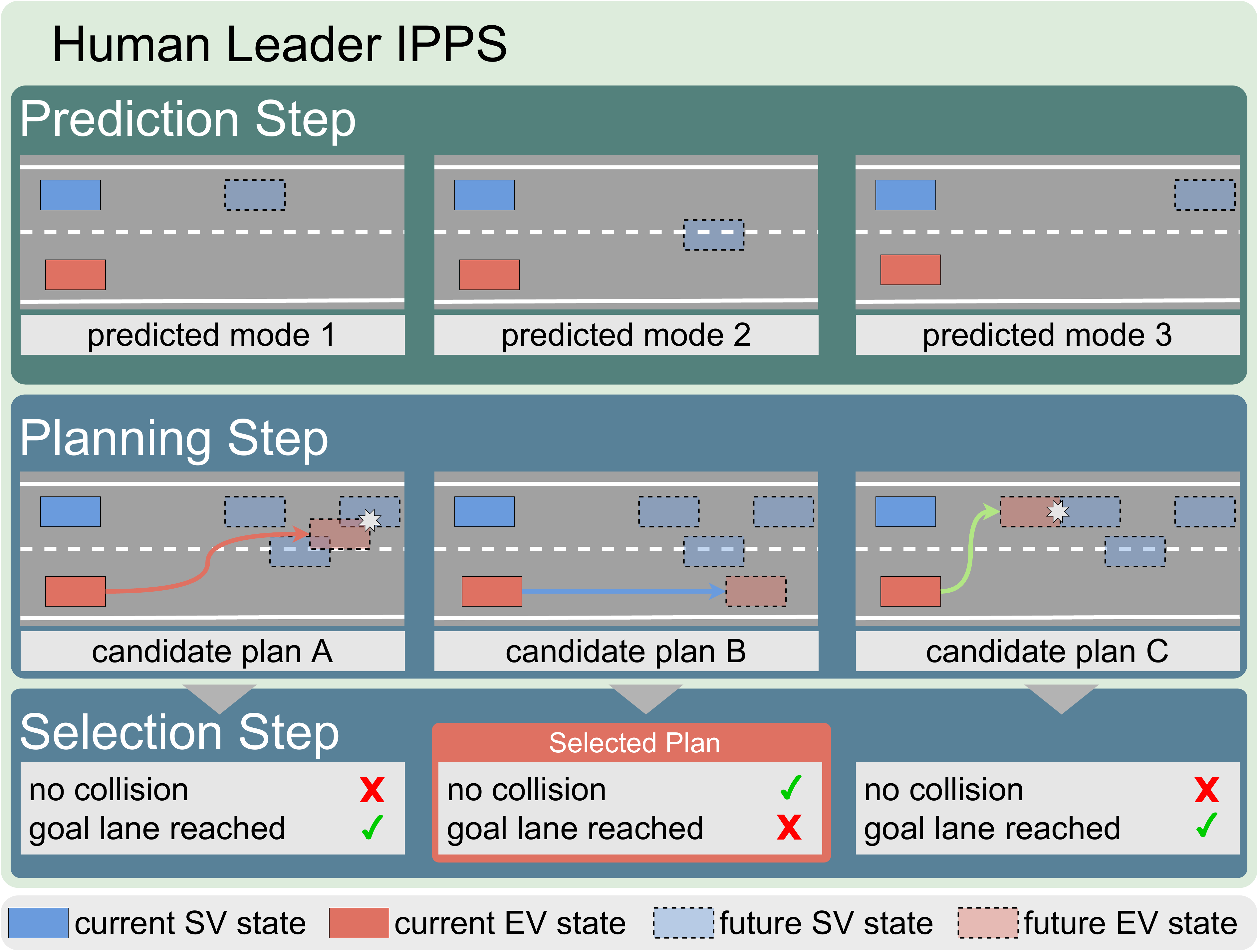}
    \caption{An exemplary sequential IPPS that falls in the human leader category. First, the SV's future is predicted, and then a plan is made based on the anticipated future. In this example, the planning step involves evaluating three candidate plans and selecting the one that does not collide with the predicted SV positions.}
    \label{fig:human_leader_ipps}
\end{figure}
Most sequential IPPSs belong to the human leader integration principle~\cite{zeng2019end,rhinehart2018deep,casas2021mp3,sadat2020perceive,zeng2020dsdnet,cui2021lookout,vitelli2022safetynet,hu2023planning,ye2023fusionad}.
Human leader IPPSs can only be realized in modular or modular E2E automated driving systems since the relation of prediction and planning is specifically engineered.
The planned EV trajectory is conditioned on the predicted SV behavior as depicted exemplarily in Fig.~\ref{fig:human_leader_ipps}.
This means the influence of the EV's plan on the SVs is not modeled.
Due to this unidirectional dependency, the EV shows a reactive behavior, which can lead to underconfident plans~\cite{rhinehart2021contingencies}.
In the lane change example in Fig.~\ref{fig:human_leader_ipps}, the EV tries to find a plan that is suitable given the three predicted SV behaviors without being aware that it can influence them.
Consequently, it will favor the conservative plan and fail to reach its goal lane.
If the EV could reason about the influence of its own actions on the predicted SV behavior, it could anticipate that the SV might slow down or speed up further in reaction to its behavior, which would make the lane change maneuver possible.
This simple example demonstrates the limitations of unidirectional, sequential IPPSs and stresses the advantage of predictions that depend on the EV's behavior.

There is a vast body of literature on models that use a human leader sequential IPPSs.
Deep structured self-driving Network (DSDNet)~\cite{zeng2020dsdnet} uses a modular E2E system architecture.
Each module builds on the outputs of the preceding module and can additionally access the latent feature of the perception backbone.
Within the IPPS, a set of potential future trajectories is predicted for each detected vehicle.
The planning module then samples candidate EV trajectories and applies a traditional hand-crafted cost function.
The cost includes the collision probability based on the SV predictions.
In contrast, the Neural Motion Planner (NMP)~\cite{zeng2019end} is based on a learned cost function. Features of a shared backbone network are fed to a prediction module that forecasts future bounding boxes and a planning module that generates a spatio-temporal cost function.
As in DSDNet, planning involves sampling feasible trajectories, evaluating them with the cost function, and selecting the best one.
Beside P3~\cite{sadat2020perceive}, LookOut~\cite{cui2021lookout}, and MP3~\cite{casas2021mp3} are further examples of this development.
Specifically, P3 and MP3 predict an occupancy map while LookOut~\cite{cui2021lookout} predicts discrete trajectories.
Then a cost function that combines rule-based and learned components is used to evaluate pre-defined candidate plans.
In comparison, SafetyNet~\cite{vitelli2022safetynet} follows a unique approach.
A monolithic E2E planner with implicit prediction (cf. Sec.~\ref{sec:implicit_ipps}) is applied in parallel to a module that outputs explicit predictions.
Collision checks between the EV plan and SV predictions are performed to ensure the plan's safety.
If the initial plan is classified as unsafe, a hand-crafted fallback layer generates a lane-aligned trajectory.
Hence, the prediction module affects the selected plan, and we classify SafetyNet as a human leader IPPS.
UniAD~\cite{hu2023planning} implements a completely learning-based modular E2E system, which employs queries as the interface between the submodules. Thus the planning module can attend to agent-level features from previous network layers.
Various works follow a similar approach: Vectorized Autonomous Driving (VAD)~\cite{jiang2023vad} and Hydra-MDP~\cite{li2024hydra} use the output features of previous layers explicitly for collision and map compliance checks during training.

\subsubsection{Robot Leader}
\begin{figure}
    \centering
    \includegraphics[width=\linewidth]{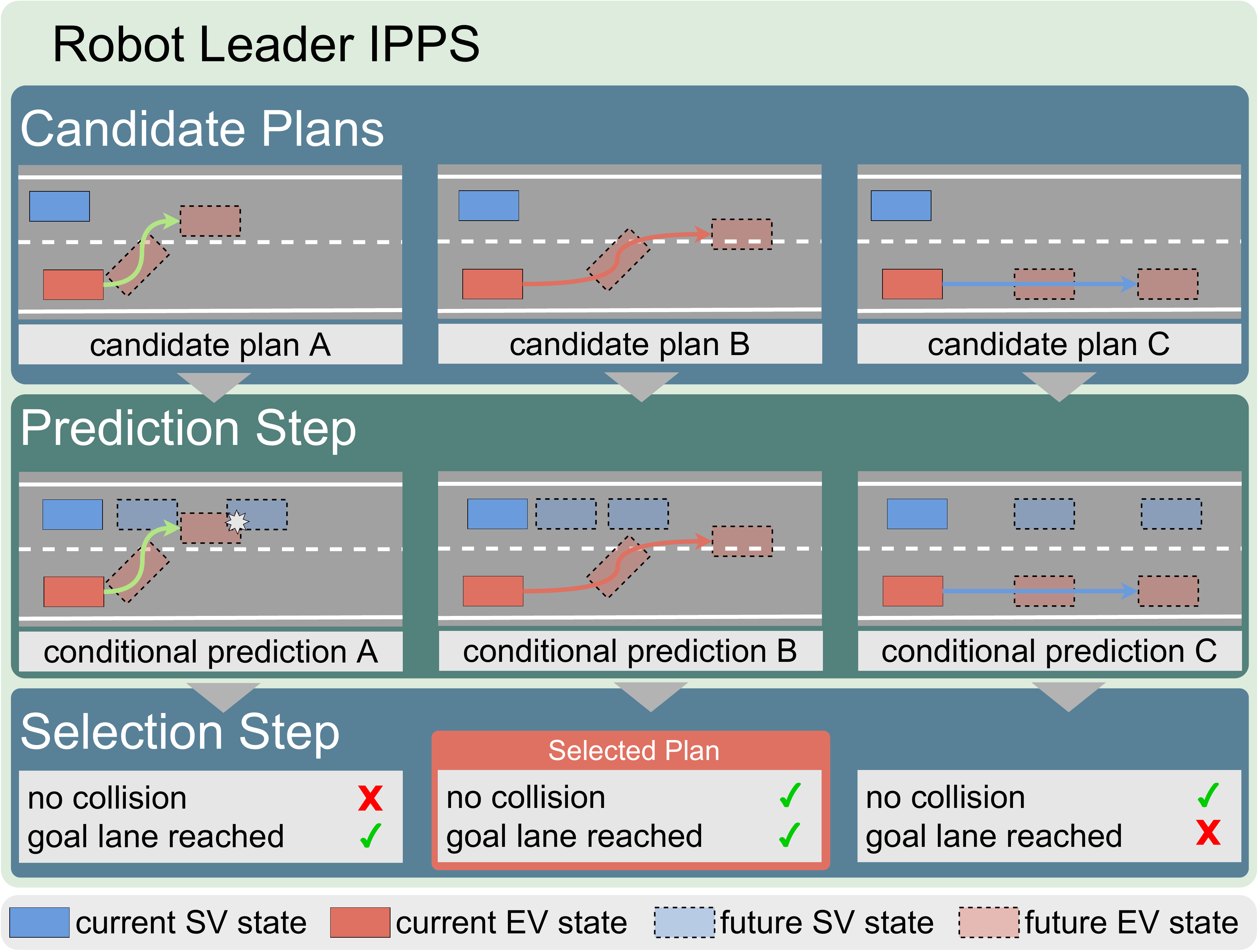}
    \caption{An exemplary robot leader IPPS. First, three candidate plans are generated, e.g., with a kinematic sampler. Then, a prediction conditioned on each candidate plan is inferred. In the example, the EV expects the SV to brake for it if it slowly merges to the left lane. Thus, it opts for this plan. Selecting this plan without hedging against the risk that the EV does not brake as expected results in overconfident behavior.}
    \label{fig:robot_leader_ipps}
\end{figure}
In robot leader IPPSs, candidate EV plans are inferred based on the current state, and the prediction step for the whole environment is conditioned on it~\cite{song2020pip}.
For instance, \cite{bae2022lane,sheng2022cooperation,song2020pip} predict SV trajectories conditioned on candidate plans for the EV with a neural module and then select the best one with a hand-crafted cost function.
Intuitively, these methods forecast different future scenarios based on pre-defined candidate plans.
Like human leader IPPSs, they can only be realized in modular or modular E2E automated driving systems due to the manually engineered influence of EV and SVs.
In contrast to human leader IPPSs, the EV can anticipate the reaction of SVs to its own actions in robot leader IPPSs.
Hence, the EV can effectively seek to make the SV react to its trajectory, which can result in aggressive driving behavior~\cite{sadigh2016planning, schmerling2018multimodal, tang2019multiple}.
Consequently, in the example in Fig.~\ref{fig:robot_leader_ipps}, the EV might reason that when it follows the fast-progressing plan B, the observed SV will slow down to prevent a collision.

Overall, robot leader IPPSs are no common choice in AD (cf. Fig.~\ref{fig:timeline}).
In a pioneering work~\cite{sadigh2016planning}, classic optimization is mixed with learned elements to model the effect of EV behavior on SVs.
The first fully learning-based method applies a conditional variational autoencoder to obtain diverse rollouts of the future of the scene recurrently~\cite{schmerling2018multimodal}.
At each timestep of the rollout, the prediction is conditioned on the next step of a pre-defined candidate plan.
Evaluating multimodal predictions for each candidate plan with a hand-crafted cost function yields the best plan.

The example in Fig.~\ref{fig:robot_leader_ipps} reveals that robot leader systems can, other than human leader systems, anticipate the reaction of SVs to EV actions.
However, this paradigm might lead to overconfident behavior.
Altogether, this section has shown that no sequential IPPS can account for the bidirectional influence of EV behavior on SVs and vice versa.
To resolve the issue, other integration principles are necessary.

\section{Undirected IPPS}\label{sec:undirected_ipps}
In contrast to sequential IPPSs, which are separated into various submodules, monolithic E2E systems employ a single neural network.
Implicit systems directly map the state inputs $\overline{\textbf{X}}_\textrm{SV}$, $X_\textrm{EV}$, and $I$ to the planned EV trajectory $Y_{\textrm{EV}}$~\cite{muller2005off}.
Alternatively, joint optimization methods perform prediction and planning in a single step.
Alongside $Y_{\textrm{EV}}$ they output SV predictions $\overline{\textbf{X}}_\textrm{SV}$.
In the following, we first examine implicit systems and then discuss joint optimization approaches.

\subsection{Implicit}
\label{sec:implicit_ipps}
Reasoning about the future outcome of the environment is crucial when planning the behavior of the EV. Hence, it can be assumed that even models without a dedicated prediction output anticipate the behavior of their environment.
Generally, the expert's actions, which the E2E planner learns to imitate in training, are based on reasoning of how the future will unfold.
By learning to imitate actions, the model could exhibit behavior that is implicitly based on such reasoning as well.
For instance, consider a scenario where the EV's leading vehicle brakes.
An expert or a human driver will base their driving decision on the likely future scenario that the safety margin will shrink and thus brake as well.
Hence, an E2E planner that imitates the expert's behavior makes a decision that is implicitly based on a predicted future.
For instance, HydraMDP~\cite{li2024hydra} is a monolithic system that plans by regressing a score for each candidate trajectory within a predefined set. In training, an auxiliary collision cost is predicted for each candidate. Predicting this accurately is impossible without reasoning about the future behavior of SVs. Thus, the SVs' future behavior and the interactions among them, as well as between the SVs and the EV, are modeled implicitly.

Pioneered by ALVINN~\cite{pomerleau1988alvinn}, there is a vast body of literature on IPPSs based on monolithic E2E planners~\cite{bojarski2016end,xu2017end,codevilla2018end,hecker2018end,muller2018driving,rhinehart2018deep,wang2019deep,chen2019deep,codevilla2019exploring,hawke2020urban,chen2020learning,ohn2020learning,prakash2021multi,scheel2022urban,li2018rethinking}.
The main disadvantage is their black-box nature, which makes model introspection and safety verification difficult.
Moreover, understanding the limitations and their capabilities with respect to their interactive behavior is difficult. In fact, this can only be evaluated empirically in comprehensive benchmarks, which is what we outline as a direction for future research in Sec.~\ref{subsec:interactive_benchmarks}.
To improve interpretability, some works generate interpretable intermediate outputs such as online HD maps~\cite{zhao2021sam}, semantic segmentations~\cite{chitta2021neat}, and object detections~\cite{chitta2022transfuser}.
These provide additional training supervision and introspection into the perception part of the system. We clearly separate this from systems that additionally output predictions. 
For instance, if the prediction output is branched off from an intermediate feature layer, the downstream part is focused exclusively on planning~\cite{bansal2018chauffeurnet, zeng2019end, chitta2021neat, renz2022plant}, while the upstream part is relevant to all subtasks. 
Given that this makes planning a downstream task of prediction, the architecture would fall into the human leader category.
In contrast, if prediction and planning are both outputs of the final layer, we assume that both tasks are addressed in a single step, which means that the architecture resembles a joint optimization system, discussed in the next section.

\subsection{Joint Optimization}
\label{subsec:joint_optimization}
\begin{figure}
    \centering
    \includegraphics[width=\linewidth]{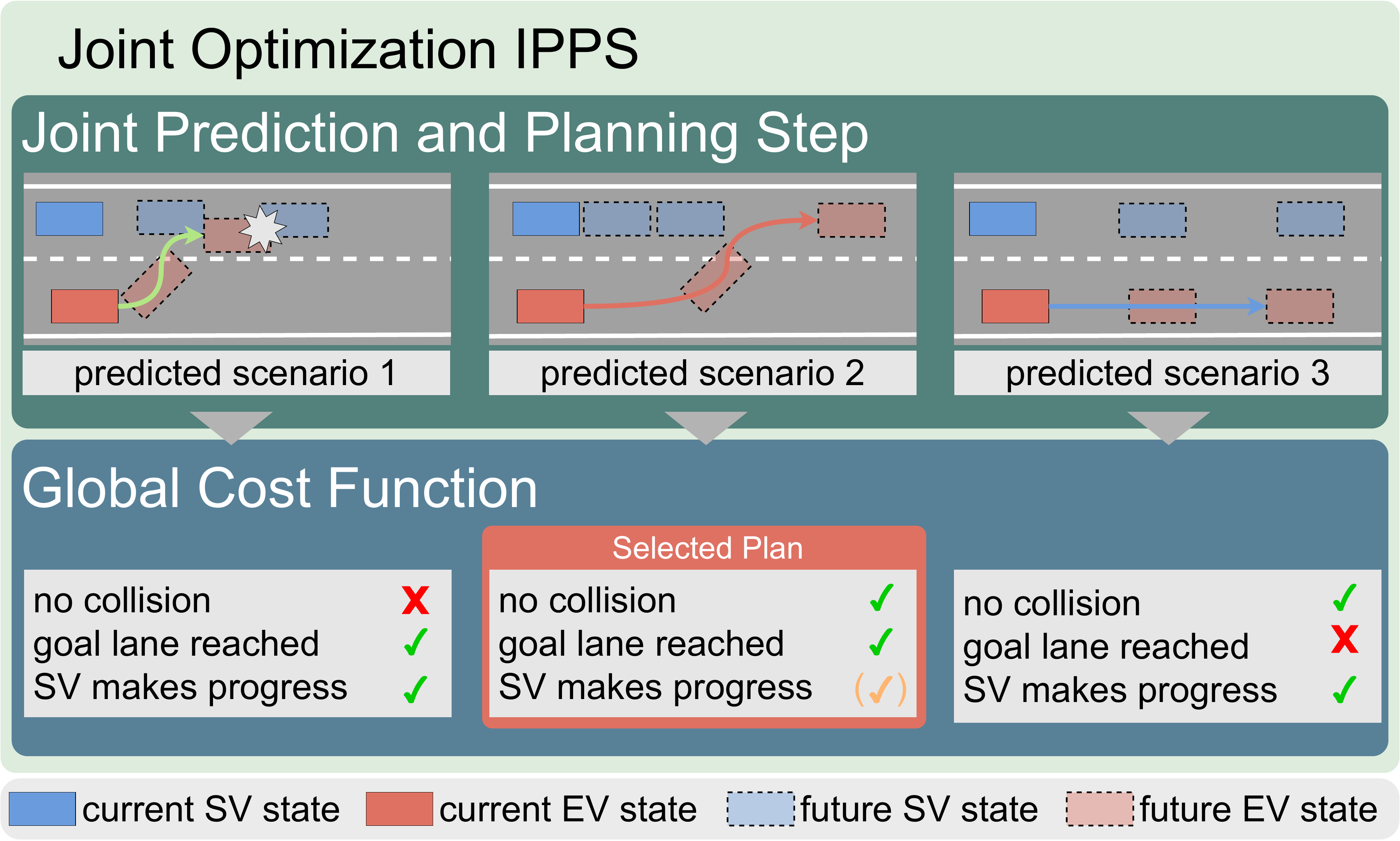}
    \caption{An exemplary joint optimization IPPS. A joint global cost function is optimized by selecting the best of several potential scenarios. Therefore, a joint prediction and planning step forecasts several potential scenarios that describe the EV and SV behavior. Then, each of them is evaluated using a global cost function that evaluates the outcome for all vehicles. Here, a plan for the EV is selected that corresponds to the scenario where the SV brakes to let the EV merge before it.}
    \label{fig:joint_leader_ipps}
\end{figure}
Joint optimization describes IPPSs that deterministically approximate a joint objective~\cite{liu2021deep} under the assumption that an optimal outcome exists.
The EV's plan is obtained by global optimization across all agents~\cite{chen2023interactive}.
This erroneously implies that the behavior model used for SVs is exactly known and that every traffic participant optimizes this same global objective.
This becomes particularly challenging when considering goal-conditioning for the EV. If the joint prediction and planning process is conditioned on the EV's goal, it suggests that all SVs will behave in a manner that enables the EV to achieve its goal.
At the architecture level, these systems predict and plan in a joint step.
This can mean explicitly optimizing a learned or hand-crafted cost function or decoding a joint prediction for all agents, including the EV. In the latter case, the optimization can be an iterative process~\cite{chen2023deepemplanner,huang2023gameformer} or limited to one step~\cite{pini2022safe}.
An exemplary architecture is shown in Fig.~\ref{fig:joint_leader_ipps}, where a global cost function is used to identify the optimal plan for the EV from a set of possible future scenarios. The selected plan relies on the SV to break so the EV can merge. 
However, the EV selects this plan without considering that the SV might not behave accordingly.
In particular, joint optimization applies to monolithic planners that map all inputs to a plan for the EV $Y_{\textrm{EV}}$ and the predicted future of the SVs $\overline{\textbf{X}}_\textrm{SV}$.
As only the EV plan is propagated to a downstream controller and the predictions are not used further, they act as a regularization which can make the learning process more sample-efficient and can improve generalization capabilities~\cite{bansal2018chauffeurnet, chen2022learning, sauer2018conditional} as it was previously demonstrated for other auxiliary tasks in prediction and planning like and occupancy forecasting~\cite{sadat2020perceive}.

A seminal work in this area is GameFormer~\cite{huang2023gameformer}.
Inspired by hierarchical game-theoretic frameworks~\cite{hang2021cooperative, wang2022social, li2017game,geiger2021learning}, it models interaction between agents as a level-$k$ game~\cite{wright2010beyond, costa2009comparing}.
Thus, a Transformer decoder models the interaction between all agents, including the EV, by iteratively updating the individual predictions based on the predicted behavior of all agents from the previous level.
Similarly, SafePathNet\cite{pini2022safe} employs a Transformer module for joint prediction and planning of multiple scenarios and then ranks them based on their estimated probability and collision avoidance.
Generative end-to-end autonomous driving (GenAD)~\cite{zheng2024genad} adapts a generative decoder to regress SV and EV trajectories jointly.
The main problem with this approach is that there is no guarantee that the SVs will behave accordingly.
Demonstrations show how this can lead to fatal errors~\cite{rhinehart2021contingencies}.
However, in applications with vehicle-to-vehicle or vehicle-to-infrastructure communication, intentions can be communicated between agents, and global optimization is a valid modeling assumption. There, a global cost function reflecting each actor's individual goals is optimized either by a central server or by negotiation among agents. The resulting plan is assigned to each vehicle leading to more efficient and safe traffic~\cite{klimke2022cooperative,klimke2023automatic}.

\section{Bidirectional IPPS}\label{sec:bidirectional_ipps}
\begin{figure*}
    \centering
    \includegraphics[width=\textwidth]{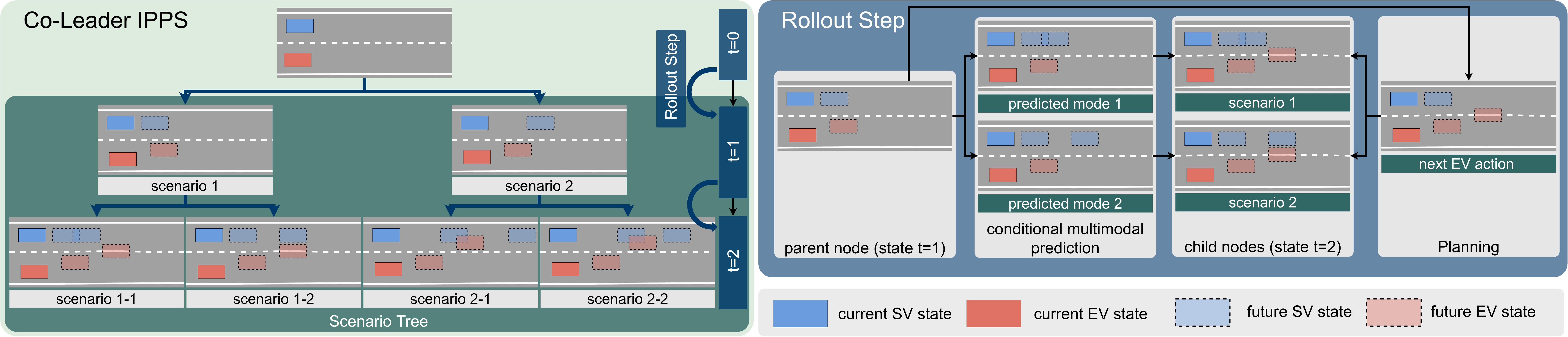}
    \caption{
    \textbf{Exemplary tree-based co-leader planner.}
    Starting at an initial state ($t=0$) a tree is constructed to model potential EV plans and SV behaviors.
    The tree spans across several stages, each reflecting a discrete timestep within the planning horizon.
    In each stage, child nodes are generated by predicting the SVs and unrolling one EV action, both conditioned on the state of the parent node. This is visualized in the ``rollout step'' section.
    Note that the state in the parent node ($t=1$) includes how far the EV has progressed along the candidate plan until this point. Thus, the prediction can only observe the candidate plan up to $t=1$. It is not conditioned on the EV's next action, i.e., it's causal.
    Due to space limitations, we only visualize a prediction with two modes and only a single EV plan per rollout step. In practice, a tree might comprise more predictions and evaluate multiple potential EV behaviors.
    Based on the anticipated outcome of each EV behavior, i.e. the evaluation of each node w.r.t. collisions, progress and comfort, the best action can be chosen by the planner.}
    \label{fig:tree_co_leader}
    \padfigure{}
\end{figure*}
We term systems that react to anticipated behaviors of SVs but also expect SVs to react to their own actions bidirectional IPPSs, as they model the influence of prediction on planning and vice versa.
Intuitively, this causes a chicken-and-egg problem, where predictions are needed to plan an appropriate behavior, and at the same time, the EV's future plan is needed to predict the SVs' reactions to it.
In this section, we highlight how such systems can be realized by predicting a reactive policy instead of a fixed behavior for SVs. 
The resulting interactive behavior is termed co-leader.

\subsection{Co-leader}
Co-leader planning sets the highest requirements for an IPPS.
It must model the influence of SVs' potential future behaviors on the EV plan as well as their stochastic reaction to potential ego trajectories~\cite{galceran2017multipolicy, fisac2019hierarchical, rhinehart2021contingencies}.
The result is an interaction-aware system that can model the anticipated behavior of its surroundings and how it can influence it. However, it does not assume this reaction to be deterministic, which is in contrast to joint optimization. Thus, the EV cannot be certain about SVs' reactions and must be prepared to react to multiple future outcomes~\cite{cunningham2015mpdm}.
For instance, in the opening example in Fig.~\ref{fig:teaser_figure}, the EV can try to approach the intersection to provoke a reaction of the SV. This reaction can either indicate that the SV is going to make a turn, e.g., with a turn signal or by slowing down, or the opposite, i.e., accelerating or keeping its speed, to indicate that the SV will continue straight.
The former will enable the EV to confidently decide in favor of the faster-progressing plan.
Even though this paradigm is most capable in theory, only a few works exist that fulfill the challenging requirements set to the system.

\subsubsection{IPPS Architectures}
Contingencies from Observations (CfO)~\cite{rhinehart2021contingencies} originally introduced this paradigm, leveraging normalizing flows to represent multi-agent behavior as a learned coupling of independent variables. However, the intricate nature of this architecture may pose obstacles to wider adoption. Furthermore, their benchmark is simplistic, encompassing only a few scenarios with hard-coded agent behaviors.
Alternatively, tree-structured methods overcome the circular dependence of prediction and planning by representing future outcomes in a tree, where each node is a state of the EV and its surroundings. 
Multimodality caused by different EV decisions or multiple potential reactions of SVs is handled by branching~\cite{zhang2020efficient, ding2021epsilon}.
Tree-structured policy planning (TPP)~\cite{chen2023tree} presents a readily comprehensible tree-structured framework that implements co-leader planning. It involves sampling kinematically feasible trajectories for the EV and then conditioning prediction on each of them. An autoregressive formulation for prediction ensures causality: it can be seen as a way to roll out the scene with SV predictions and the EV plan such that both can only access the observation history up until the respective rollout step.
Autoregressive decoding can help to enforce high-quality social interactions and scene understanding~\cite{rhinehart2018r2p2, ivanovic2019trajectron, salzmann2020trajectron++, liu2021multimodal, yuan2021agentformer, cui2022gorela} but might lead to compounding errors since predicted information is used to make further predictions~\cite{ross2011reduction}.
Finally, the best EV trajectory proposal is selected with a dynamic programming algorithm.
Differentiable tree-structured policy planning (DTPP)~\cite{huang2023dtpp} extends this with a Transformer-based conditional prediction module, which allows the forecast of all agents conditioned on multiple EV plans at once. 
It also replaced the dynamic program with a learnable cost function and applied node pruning to reduce computational burden. Still, the EV reaction to the prediction is taken over entirely by the cost function because the sampled ego-proposals are non-reactive with respect to the SV predictions.
In general, representing the potential EV and SV behaviors in a temporal tree is easy to implement and interpret. 
An exemplary tree-based co-leader planning architecture similar to TPP and DTPP is shown in Fig~\ref{fig:tree_co_leader}. 
It can be seen that the future planning horizon has to be discretized into a limited number of steps that refer to the levels of the tree.
Further, the figure shows, how the number of leaf nodes grows exponentially with the branching factor, the depth and the number of candidate plans.
The branching factor has to be large enough to accurately reflect the uncertainty in the SVs' behavior as well as the behavior options of the EV and the number of candidate plans can be very high if a dense kinematic sampler is used.
Moreover, discretizing the future horizon over which a scenario unfolds introduces a sensitive hyperparameter. The discretization timestep is subject to a tradeoff between runtime and accuracy w.r.t. capturing all relevant intermediate states.
To counter this, TPP~\cite{chen2023tree} and DTPP~\cite{huang2023dtpp} build a tree with very limited depth of 2 stages. TPP uses a branching factor of 4, while DTPP forecasts 5 states in the first stage and branches each into 6 in the second stage.
This makes architectures, which are able to model how the future unrolls without relying on discrete timesteps a promising opportunity for future research.

\subsubsection{Scenario-Based Trajectory Planning}
Bidirectional IPPSs, i.e., co-leader planning methods, are more complex than sequential or undirected IPPSs because they anticipate and consider multiple future scenarios and to make plans based on this. For instance, sequential IPPSs can be used with unimodal predictions or employ a single representation for all scenarios (e.g., occupancy grids), whereas co-leader planning methods must model the multimodal stochastic behavior of their surroundings and plan accordingly. 
Similarly, monolithic E2E IPSSs with undirected integration (cf. Sec.~\ref{sec:undirected_ipps}) do not explicitly consider multiple future scenarios.
Hence, the representation marginalizes all future outcomes.
While this is easy to realize, it can be safety-critical. In the example of Fig.~\ref{fig:teaser_figure}, the cost function must balance high progress and the potential safety risk if the unlikely prediction occurs for one plan with low progress for the other.
Hence, it must trade off unlikely but dangerous scenarios (e.g.\ collision) against likely low-cost scenarios.
Conversely, a co-leader architecture must make plans in the face of the uncertain reaction of SVs to its plans.
In the following, we will discuss methods that incorporate multiple outcome scenarios when choosing among various EV plans.
We denote the number of relevant future scenarios with $N_s$.
They can be reflected by a multinomial distribution $P\left(\overline{\textbf{Y}}^{(i)}_\textrm{SV}\right), i=1,..,N_s$ describing the behavior of SVs and the EV in each scenario.
Based on this, we describe two groups of existing methods: Worst-case planning and contingency planning.
Their difference lies in the cost function component $h$ of the planning function $f=h(g(X_\textrm{EV}, \overline{\textbf{X}}_\textrm{SV}, I)$, which selects one plan among a set of potential plans generated by $g$. 
In the architecture described in Fig~\ref{fig:tree_co_leader}, $g$ is the tree generation step (green). The selection function $h$ would evaluate all nodes and select the most suitable plan (not visualized).

\textit{Worst-case planning} refers to IPPSs that evaluate each plan based on the worst-case outcome scenario.
In the tree-based example, this would mean that if there is a collision in one node, then all nodes that share the same action would be discarded.
No probabilities $P(\overline{\textbf{Y}}^{(j)}_\textrm{SV})$ are considered.
The motivation behind this assumption is that the final output trajectory must be safe in each possible scenario.
Consequently, the selection function $h$ can be formulated as
\begin{equation}
    h=\argmin_i \max_j c(Y^{(i)}_\textrm{EV},\overline{\textbf{Y}}^{(j)}_\textrm{SV}).
\end{equation}
This paradigm strongly focuses on collision avoidance and can result in overly conservative behavior, as shown in Fig.~\ref{fig:teaser_figure}, where only the slower progressing plan is safe under all predictions. Thus, it is preferred under worst-case assumptions.

\textit{Contingency planning} aims to improve this by additionally taking into account the probabilities $P(\overline{\textbf{Y}}^{(j)}_\textrm{SV})$ of different future scenarios $\overline{\textbf{Y}}^{(j)}_\textrm{SV}$~\cite{hardy2013contingency, li2023marc, li2024multi}. 
The resulting plan hedges against worst-case risks while enabling progress in expectation.
For instance, \cite{cui2021lookout, hardy2013contingency, zhan2016non} find a plan that is safe for scenarios on the short horizon (independently of their probability, i.e., worst-case) and also optimize the expected cost on the long horizon (marginalization).
Thus,
\begin{equation}
    h=\argmin_i \left( \max_j c_{s}(Y^{(i)}_\textrm{EV},\overline{\textbf{Y}}^{(j)}_\textrm{SV}) + \mathbf{E}_j[c_{l}(Y^{(i)}_\textrm{EV},\overline{\textbf{Y}}^{(j)}_\textrm{SV})] \right),
\end{equation} where $c_s$ and $c_l$ consider the short-term and the long-term part of the proposal $Y^{(i)}_\textrm{EV}$, respectively.
In the example in Fig.~\ref{fig:teaser_figure}, this planner allows the most progress while still ensuring safety.
It can exploit that both candidate plans have a common short-term action. Hence, it could opt for the faster-progressing plan as it is aware that it could still re-plan and fall back to the slower one.
This method can also be applied to non-bidirectional IPPSs. For instance, it can be combined with a sequential human leader IPPS, which forecasts multiple scenarios.
CfO~\cite{rhinehart2021contingencies} terms this \textit{passive contingency}, as the EV is contingent upon the SVs~\cite{cui2021lookout, hardy2013contingency, zhan2016non, bajcsy2021analyzing}. In contrast, \textit{active contingency} refers to co-leader systems that are aware that the EV can influence the SVs' behavior behavior~\cite{galceran2017multipolicy, fisac2019hierarchical, bandyopadhyay2013intention}.

\section{Discussion and Frontiers}
\label{sec:discussion_frontiers}

\subsection{Categorization Edge-cases}
In the previous sections, we described and discussed different system architectures and paradigms to integrate prediction and planning in interaction-aware autonomous systems. 
Here, we want to emphasize the distinction between these categories and describe corner cases that transition fluently from one category to another with minor architectural tweaks.
For instance, an IPPS that directly maps observational inputs to a trajectory for the EV is clearly a monolithic IPPS that falls into the implicit integration category. 
Extending it with an additional decoder head that outputs predicted trajectories for SVs makes it more interpretable. For instance, NMP~\cite{zeng2019end} employs such an architecture.
Thus, the changed architecture consists of a shared backbone for all tasks, which is connected to a prediction decoder module as well as a downstream planner module.
Therefore, this change transforms the monolithic IPPS into a modular system. As the planning step comes after prediction, the system falls into the human leader category.
Similarly, extending the implicit IPPS architecture with a prediction output from the final layer means that the entire network performs prediction and planning in a joint step, turning it into a joint optimization IPPS. 
Similarly, the exemplary architecture presented in Fig.~\ref{fig:robot_leader_ipps} can be modified: It is a robot leader IPPS because it expects the SV to react deterministically to the EV's actions, and the plan is selected with only the EV's goals in mind.
However, if the cost function used in the final selection step also optimizes the SV's outcome, the architecture would fall into the ``joint optimization'' category.

\subsection{Benchmarking IPPS}
In general, IPPSs need to be evaluated as a system, either because no submodules can be distinguished (monolithic E2E) or because of the interdependence of the submodules.
\subsubsection{Prediction Benchmarks}
A special case are widely adopted sequential human leader architectures, where the prediction component can be evaluated independently of the downstream planning task.
Various prediction datasets provide expert recordings split into training, validation, and test sets for benchmarking.
The most prominent ones are nuScenes~\cite{ caesar2020nuscenes}, WOMD~\cite{ettinger2021large}, Argoverse~\cite{chang2019argoverse}, Argoverse2~\cite{wilson2023argoverse}, inD~\cite{bock2020ind}, highD~\cite{krajewski2018highd}, rounD~\cite{krajewski2020round}, exiD~\cite{exiDdataset}, and Interaction~\cite{zhan2019interaction}.
However, optimizing the prediction performance might not lead to better planning. In fact, some works even found that optimizing prediction metrics leads to worse planning results and, thus, worse driving performance~\cite{huang2023dtpp}.
One reason for this is that prediction metrics average across all target vehicles, i.e., a prediction error for a vehicle far away has the same impact on the score as an error for a vehicle right in front of the EV. 
Moreover, the isolated assessment of prediction modules does not consider the interplay with other submodules. For instance, a planner might navigate the EV to states that cause severe distributional shifts and impair prediction performance.
Thus, it is broadly acknowledged that IPPSs must be evaluated as a system.

\subsubsection{Open-loop Evaluation} 
There are two methods to evaluate ADSs as a whole, namely open-loop evaluation and closed-loop simulation.
Open-loop evaluation is similar to an evaluation in the prediction task.
It compares the planner output $Y_\textrm{EV}$ to that of an expert planner $Y_\textrm{GT}$~\cite{cheng2023rethinking}. 
However, this does not give the planner control over the EV and ignores the problems arising from compounding errors and distributional shifts~\cite{li2024ego}.
More severely, recent work demonstrated that open-loop evaluation results do not correlate with driving performance~\cite{dauner2023parting,codevilla2018offline}.

\subsubsection{Closed-loop Simulation}
Instead of merely comparing the planner output $Y_\textrm{EV}$ to that of an expert planner $Y_\textrm{GT}$, closed-loop simulations let the planner control the EV. The evaluation is based on metrics for comfort, safety, and progress.
Thus, closed-loop evaluation relates better to real-world driving but requires a simulator to let the model interact with its environment.
Prominent simulators are Carla~\cite{dosovitskiy2017carla}, nuPlan~\cite{caesar2021nuplan}, Waymax~\cite{gulino2024waymax}, Highway-Env~\cite{leurent2018environment}, and CommonRoad~\cite{althoff2017commonroad}.
Carla is a high-fidelity simulator that involves sensor simulation, whereas nuPlan, Waymax, CommonRoad, and Highway-Env operate on abstract object representations (e.g., bounding boxes) without rendering camera or lidar sensors.
Many benchmark datasets were proposed for the Carla simulator~\cite{codevilla2018offline, prakash2021multi, zhang2021end, chitta2022transfuser}.
However, the scenarios are purely synthetic, which questions the generalizability to the real world.
In contrast, nuPlan and Waymax follow a data-driven approach, which uses recordings from real-world scenarios to initialize the simulation. Thus, users have access to corresponding large-scale real-world datasets to train their method before deploying it in simulation.
CommonRoad is compatible with several smaller datasets~\cite{bock2020ind,krajewski2018highd,krajewski2020round,exiDdataset,zhan2019interaction}. Highway-env is based on simplistic synthetic environments.

\subsubsection{Long-tail Scenarios}
When simulating regular driving scenes, rare and critical scenarios are naturally under-represented.
This problem can be tackled with data curation~\cite{filos2020can, wu2023policy} or additional test case generation.
These can be generated manually~\cite{kluck2023empirical, rocklage2017automated, birkemeyer2022feature, mi2021hdmapgen, erz2022towards,hallgarten2024can} or automatically from recorded driving logs~\cite{gambi2022generating, langner2018estimating, hauer2020clustering}.
Other methods generate adversarial critical scenarios by augmenting actors within the scene~\cite{wang2021advsim, abeysirigoonawardena2019generating, althoff2018automatic, indaheng2021scenario, klischat2019generating, kothari2021human, wachi2019failure,hanselmann2022king} or the scene itself~\cite{zhou2020deepbillboard, kong2020physgan, boloor2020attacking, sato2021dirty, bahari2022vehicle}.
Similarly, \cite{rhinehart2021contingencies, codevilla2019exploring} investigate human-robot interaction more closely with hand-crafted scenarios – an aspect that is uncared for in average driving situations.

\subsubsection{Traffic Agent Simulation}
Another major challenge in closed-loop evaluation is to realistically simulate the SVs.  
In NAVSIM~\cite{dauner2024navsim}, Waymax, and the non-reactive version of nuPlan, traffic agents follow predefined trajectories obtained from real-world recordings.
However, if the EV behaves differently than the expert during recording, this can result in collisions caused by the SVs, as they do not react to the EV.
Hence, this simulation is usually limited to short sequences, where the EV stays close to the expert's trajectory~\cite{zhang2022rethinking}.
Carla, nuPlan, and Waymax provide a reactive driver model for the SV, such as the Intelligent Driver Model (IDM)~\cite{treiber2000congested,kesting2010enhanced,derbel2013modified,albeaik2022limitations,sharath2020enhanced}.
In highway-env, traffic agents can be simulated with a driving model based on IDM+MOBIL~\cite{kesting2007general}, which is capable of performing lane changes.
Another line of work focuses on learning-based methods to simulate realistic driving behavior~\cite{suo2021trafficsim,mixsim2023,igl2022symphony,xu2022bits,guo2023scenedm,zhong2023guided}.
In Waymax~\cite{gulino2024waymax}, users can plug in their own model to control objects in the scene.

\subsubsection{Discussion}
Despite recent advancements in scenario generation and traffic simulation, there are no prominent large-scale benchmarks to evaluate the interactive behavior of IPPSs.
Real-world datasets commonly comprise a limited number of lane changes or unprotected turns, and synthetic datasets with hand-crafted scenarios don't scale to the needs of comprehensive benchmarking. Moreover, leading benchmarks do not report metrics highlighting the strengths and weaknesses of different IPPS architectures concerning their interactive behavior. We emphasize this as an important avenue for future research and hope our work inspires researchers to tackle this problem.
\begin{figure*}
    \centering
    \resizebox{1.0\linewidth}{!}{\input{images/timeline.pgf}}
    \caption{Important contributions to DL-based planning. We highlight how prediction and planning are integrated. The nodes and stem of each entry encode different aspects, which are explained below the timeline. To avoid overlaps works marked with $\leftarrow$ and $\rightarrow$ were slightly moved into the past and future, respectively.}
    \label{fig:timeline}
\end{figure*} 

\subsection{Challenges and Trends}
\label{sec:challenges}
In the following section, we discuss existing trends and future challenges in the field of IPPSs.
Based on our overview of deep learning-based IPPSs and their components, we identify three challenges for future research: system design, comprehensive benchmarking, and vehicle-to-X communication. After discussing them, we draw a final conclusion.

\subsubsection{Trends} A chronological overview and categorization of influential works in the field is given in Fig.~\ref{fig:timeline}.
Clustering research trends based on this overview yields three main phases of IPPSs.
Until 2019 mainly monolithic systems with implicit prediction and output regression were used.
Following works since 2020 parted with regression and instead incorporated prior knowledge in interpretable cost functions. 
Most of them integrated unconditioned predictions into the cost function and thus followed a human leader approach, which can lead to overly cautious behavior.
Many works have already modularized the monolithic system and are E2E-trainable.
Notably, almost all current state-of-the-art systems are end-to-end differentiable.
In a third phase, beginning at the end of 2022, the integration of prediction and planning is getting more complex.
In addition to sequential systems with human leader integration, undirected systems with joint IPPSs and bidirectional systems with co-leader IPPSs are implemented more frequently.
Hybrid planning is also used more frequently than before.
After cost functions were initially used to evaluate fixed candidate plans, newer methods often regress the candidates based on the scene and then apply the cost function.

\subsubsection{System Design}
Our survey suggests that strictly sequential architectures cannot meet the requirements set for driving systems.
Nonetheless, it remains unclear which integration architecture is most effective and if the theoretical advantages of bidirectional IPPSs can be realized in practice. Especially the interactive behavior has seen little attention so far. We believe that E2E differentiable bidirectional IPPSs offer great potential for future study.

\subsubsection{Benchmarking in Interactive Scenarios} 
\label{subsec:interactive_benchmarks}
We discussed different aspects of various IPPS architectures and design paradigms.
However, no comprehensive empirical benchmark reproduces and analyzes their strengths and weaknesses.
Such an overview would help to better understand the effects of different system architectures.
This requires simulation in realistic and highly interactive scenarios with realistic driver models for surrounding vehicles and expressive interaction metrics.
In this regard, data-driven simulation is a fundamental direction for future research. 
While powerful generative models have already been applied to scenario generation, simulating and evaluating the interactive behavior requires appropriate reactive policies to simulate traffic agents.
Moreover, benchmarks must cover corner case scenarios and test the generalization to distributional shifts.

\subsubsection{Vehicle-to-X Communication}
The systems described in this work leverage predictions to represent the anticipated intentions of other traffic participants. With vehicle-to-vehicle or vehicle-to-X communication, these intentions don't need to be predicted but can be transmitted to surrounding agents. Thus, joint optimization methods (see Sec.~\ref{subsec:joint_optimization}) that can optimize several vehicles' goals can improve traffic efficiency and safety. However, such systems also need to hedge against erroneous transmissions or even malicious attacks. Moreover, they still need to be able to interact with non-communicating human-driven vehicles. Consequently, extending IPPS architectures to incorporate this additional information is a promising direction for future research.

\subsection{Conclusion}
In this work, we surveyed and analyzed the integration of prediction and planning methods in automated driving systems based on a comprehensive overview of the individual tasks and respective methods.
We described, proposed, and analyzed categories to compare integrated prediction and planning works and highlighted implications on safety and behavior.
Based on this, we discussed the shortcomings of undirected and sequential IPPSs and highlighted the need for IPPSs that model bidirectional interaction.
We hope that this way of approaching the architectures of IPPSs opens a new perspective to understanding their implications and capabilities with respect to interaction with traffic.
Finally, we pointed out promising directions for future research based on the identified gaps.

\ifCLASSOPTIONcaptionsoff
  \newpage
\fi

\bibliographystyle{IEEEtran}
\bibliography{literature}

\end{document}